\documentclass[runningheads]{llncs}
\usepackage[
    left = `,%
    right = ',%
    leftsub = ``,%
    rightsub = '' %
]{dirtytalk}
\usepackage{longtable}
\usepackage{multirow}
\usepackage{subcaption}
\usepackage{tabularx}   
\usepackage{url}  
\usepackage{graphicx}
\usepackage{hyperref}
\hypersetup{
    colorlinks=true,
    linkcolor=blue,
    filecolor=magenta,      
    urlcolor=cyan,
    pdftitle={Overleaf Example},
    pdfpagemode=FullScreen,
    }
\usepackage{colortbl}
\usepackage{fancybox}
\usepackage{xcolor}
\usepackage{tikz}
\newcommand\copyrightnotice[1]{
    \begin{tikzpicture}[remember picture,overlay]
    \node[anchor=south,yshift=10pt] at (current page.south) {\fbox{\parbox{\dimexpr\textwidth-\fboxsep-\fboxrule\relax}{#1}}};
    \end{tikzpicture}
}
\begin{document}

\title{Facial Analysis Systems and Down Syndrome}
%
%
 \author{Marco Rondina\inst{1}\orcidID{0000-1111-2222-3333} \and
 Fabiana Vinci \and
 Antonio Vetrò\inst{1}\orcidID{1111-2222-3333-4444} \and
 Juan Carlos De Martin\inst{1}\orcidID{2222-3333-4444-5555}}
 \institute{Politecnico di Torino, Corso Duca degli Abruzzi 24, 10129 Torino, Italy
 \email{\{marco.rondina,antonio.vetro,demartin\}@polito.it}}

 \authorrunning{M. Rondina, F. Vinci et al.}
\maketitle              
\begin{abstract}
The ethical, social and legal issues surrounding facial analysis technologies have been widely debated in recent years.
Key critics have argued that these technologies can perpetuate bias and discrimination, particularly against marginalized groups.
We contribute to this field of research by reporting on the limitations of facial analysis systems with the faces of people with Down syndrome: this particularly vulnerable group has received very little attention in the literature so far.

This study involved the creation of a specific dataset of face images.
An experimental group with faces of people with Down syndrome, and a control group with faces of people who are not affected by the syndrome.
Two commercial tools were tested on the dataset, along three tasks: gender recognition, age prediction and face labelling.

The results show an overall lower accuracy of prediction in the experimental group, and other specific patterns of performance differences: i) high error rates in gender recognition in the category of males with Down syndrome; ii) adults with Down syndrome were more often incorrectly labelled as children; iii) social stereotypes are propagated in both the control and experimental groups, with labels related to aesthetics more often associated with women, and labels related to education level and skills more often associated with men. 

These results, although limited in scope, shed new light on the biases that alter face classification when applied to faces of people with Down syndrome. 
They confirm the structural limitation of the technology, which is inherently dependent on the datasets used to train the models.
\end{abstract}
\section{Introduction and motivation}\label{sec:introduction}
\copyrightnotice{This preprint has not undergone peer review (when applicable) or any post-submission improvements or corrections. The Version of Record of this contribution is published in ``Machine Learning and Principles and Practice of Knowledge Discovery in Databases. ECML PKDD 2023. Communications in Computer and Information Science, vol 2133. Springer, Cham.'', and is available online at \href{https://doi.org/10.1007/978-3-031-74630-7_10}{https://doi.org/10.1007/978-3-031-74630-7\_10}}

In recent years, the ethical, social and legal implications of Facial Analysis Systems (FASs) arose in several parts of the world.
Several municipalities and governments banned the use of facial recognition technologies in public spaces, such as the city of San Francisco \cite{sanfranciscoboardofsupervisorsOrdinanceNo107192019}. 
In Italy, a moratorium \cite{parlamentoitalianoTestoCoordinatoDecretolegge2021} suspended the use of these systems in public spaces and by private agencies, except for activities related to criminal justice.
The European Commission proposed a restriction to 'real-time' remote biometric identification systems \cite{europeancommissionProposalRegulationEuropean2021} while the European Parliament Research Services published an analysis on the regulation of facial recognition in the EU \cite{europeanparliamentRegulatingFacialRecognition2021}.

One of the main criticisms of FASs is their potential to perpetuate biases and discrimination. 
This is particularly the case for marginalized groups such as people of colour \cite{westDiscriminatingSystemsGender2019,buolamwiniGenderShadesIntersectional2018}, or individuals with non-binary gender and transgender identities \cite{melendezUberDriverTroubles2018,vincentTransgenderYouTubersHad2017}. 
The amount of evidence and the implications of these issues are so significant that major companies have slowed development. 
Some decided to remove gender prediction from their models \cite{hillMicrosoftPlansEliminate2022} or to oppose the use of facial recognition technology for certain purposes \cite{krishnaIBMCEOLetter2019}.

This paper is part of the strand of research into the bias of FASs. 
It focuses on the limitations of FASs in relation to people with Down syndrome, an overlooked vulnerable group in this field.
We built a dataset of images of people with and without Down syndrome (200 in each group, equally divided by binary gender) and used it to test and compare the classification of two commercial tools. 
Our motivation was to provide new evidence on the structural limitations of FASs and their high dependability from training data. 
We did this from the perspective of a vulnerable social group that is often excluded from the design process and from the most popular discourses on AI bias and discrimination. 

This paper is organized as follows: in Section \ref{sec:rel_work} we position our work in relation to previous studies related to the topic of the research.
In Section \ref{sec:design}, we describe the design of our study, highlighting the research questions, the methodology used to create the test dataset and the details of the tested models.
In Section \ref{sec:results} we present the results of our experiment for the three different tasks analysed, and the related discussion.
Section \ref{sec:threats-validity-limitation} outlines the threats to validity and ethical concerns associated with the current study. 
Section \ref{sec:conclusions} summarizes the reflections on the whole experiment.
Finally, section \ref{sec:future_work} explores possible future implementations.


\section{Related Works}\label{sec:rel_work}
Several scholars highlighted the ethical issues of FASs.
Crawford in \textit{Atlas of AI} \cite{crawfordAtlasAI2021} reconstructed the history and development of AI in relationships with a variety of impacts (e.g., on the environment, work, health) and highlighted the epistemic issues of FASs and their controversial historical origins.

Other studies focused on the failures of FASs and the associated negative impacts on society. 
In terms of gender and ethnicity, Buolamwini and Gebru \cite{buolamwiniGenderShadesIntersectional2018} evaluated different commercial gender classification systems, and found that darker-skinned women were the most misclassified group.
Similar findings are reported by Klare
et al. \cite{klareFaceRecognitionPerformance2012}, who found that commercial and non-trainable algorithms performed worse for women, blacks and young people.
The issue of diversity and inclusion in FASs can arise from the lack of examples of subpopulations in the train dataset, but also from the definition of classes that incorporate specific values and beliefs, as in a binary formulation of gender \cite{scheuermanHowComputersSee2019}.
The consequences of a non-inclusive operationalization of FASs can be seen in the case of transgender representation \cite{keyesMisgenderingMachinesTrans2018}.

Recent work on the unknown behaviour of neural networks, has shown which are the key features used by commercial face classification services in order to classify gender. 
In fact, lip, eye, cheek structure and make-up are more discriminative than skin and hair length \cite{muthukumarUnderstandingUnequalGender2018}.
As the authors discuss, the fact that the make-up is so important in predicting female gender is a troubling stereotype.


Taking the next step and linking FASs with Down syndrome, several papers are based on recognizing the disability in children in their first years of life.
Agbolade et al. \cite{agboladeSyndromeFaceRecognition2020} presented a performance comparison of different machine learning methods on the task of Down Syndrome detection.
Paredes et al. \cite{paredesEmotionRecognitionSyndrome2022} compared different machine learning and deep learning techniques to perform the emotion detection task on people with Down syndrome.
Finally, Qin et al. \cite{qinAutomaticIdentificationSyndrome2020} presented an identification method based on deep convolutional neural networks. 
The studies mentioned above were aimed at identifying the syndrome through the face or better understanding emotions through technology. 
What is not covered in previous works is whether people with Down syndrome are discriminated against, in terms of lower performance of the FASs, compared to non-affected people.
This is the observable gap that we address in this paper.


\section{Study Design}\label{sec:design}
We describe the research questions that drove the entire analysis (Section \ref{sec:rq}), how the test set was constructed (Section \ref{sec:Dataset}) and how the FASs were selected (Section \ref{sec:models}).


\subsection{Research Questions}\label{sec:rq}

\textbf{RQ1: How does the FASs work, with images of Down-syndrome people, regarding the predictions of a) gender and b) age?}

Previous work has shown how FASs can fail to predict age and gender for vulnerable people (Section \ref{sec:rel_work}), but never for people with disabilities.
In this work, we are interested in understanding whether predicting gender and age for people with Down syndrome work in the same way as for people without the syndrome.
The consequences of incorrect gender and age predictions are diverse and depend on the decisions that are made according to these predictions (e.g., hiring, wrong associations for personalized contents, etc.).
\newline

\textbf{RQ2: Are the labels assigned differently to people with or without Down syndrome by image recognition models?}

The labels generated by the models and associated with an input image, can be used by companies for many purposes.
The labels are generally correlated with gender, objects in the image and emotions of the people.
The aim of this part of the study is to analyse the labels of the images and to evaluate any possible differences between people with and without Down syndrome.
These differences could lead to downstream discrimination effects, depending on the application context in which the FASs are used.

\subsection{Dataset}\label{sec:Dataset}
 In the past, some researchers have used datasets of faces of people with Down syndrome.
 They focused their experiments on classifying people with or without the syndrome \cite{agboladeSyndromeFaceRecognition2020,qinAutomaticIdentificationSyndrome2020}. 
 However, they focused exclusively on children, who are not representative of the population as a whole.

It was therefore necessary to build a set of facial images of people with Down syndrome from scratch: this set is the experimental group (EG). 
Due to a lack of resources, it wasn't possible to build a sample by taking photos directly or by contacting people and asking them to send their photos (in both cases with their explicit consent).
It was therefore necessary to use images already available on the web. 
The images come from Google searches and from websites that offer free stock images, such as iStock and Pexels. 
In this way, it is impossible to know whether the individuals have given their explicit consent: for this reason, we err on the side of caution and do not redistribute them (see the discussion in section \ref{sec:threats-validity-limitation}).
Not every EG image has a referenced age: the age is known for 66 EG males and 64 EG females. 
It is also important to note that the life expectancy of people with Down syndrome is currently around 65 years \cite{kazemiSyndromeCurrentStatus2016}.
The resulting EG set consisted of 200 images.

The control group (CG, i.e. images of people not affected by Down syndrome) also consisted of 200 images: these were collected by selecting some of the best quality images from the UTKFace dataset\footnote{\url{https://susanqq.github.io/UTKFace/}} \cite{zhangAgeProgressionRegression2017}.  
This dataset is a well-known large-scale face dataset with a long age range (from 0 to 116 years), constructed from images of famous people. 

A total of 400 images make up the dataset, 100 for each of the following categories: EG male, EG female, CG male and CG female. 
Each image in the dataset was stored in two different ways, according to the reference rule of one of the tools used (see section \ref{sec:models} that require cropped  - representing only the part of the images with the face - for gender and age detection \footnote{this is the ClarifAI model, rule number 1, described in Appendix A.2 of the Supplementary Material}).
Thus, on the one hand, the \textit{cropped} version of the images was used for gender and age recognition. 
On the other hand, the \textit{not cropped} version of the images, was used for label detection.

All images in the dataset were paired with gender and age to test the predictions of the models.
In line with the categories used by most FASs, gender is considered to be binary: we are aware of the limitations of such a representation.
Age is the corresponding age of the person at the time the photo was taken.
The information on \textit{age} was the most difficult to find. 
For the EG, we knew the age for 66 and 64 pictures of the female and male, respectively.
Instead, the CG was created using a pre-existing dataset containing at least one image for each age from 3 to 85, so that all images had the age information.
The images without age information were not used to predict age, but were used to predict gender and labels.

One of the most important aspects for good model performance is the quality of the images.
For this reason, all images were manually selected, and as a result, the majority of samples in the dataset meet the Pose, Illumination and Expression (PIE) rules \cite{phillipsIntroductionGoodBad2011}.

\subsection{Models}\label{sec:models}
The selection of services offering facial analysis, was based on a study of the most widely used and well-known commercial services, that meet some initial constraints:
\begin{itemize}
    \item the images should not be retained for use for other purposes, or at a minimum be deleted with account suspension;
    \item it should be possible to predict gender and age;
    \item there should be a free amount of test operations.
\end{itemize}
The two services that met the previous conditions are: ClarifAI and AWS Rekognition \footnote{regarding the first requirement, please refer to the policies on Appendix A in the Supplementary Material, in particular AWSR rule number 4, ClarifAI rule number 2}.
Some other services were considered, but not selected because they did not meet the above conditions.
In particular: the services retain images indefinitely, as in the case of Face++ and Mega Matcher; they require a payment for the operations, Face++ and Cognitec's Face VACS; they do not provide gender recognition, Microsoft Face API, or they are deprecated, IBM Watson Visual Recognition. 

Both ClarifAI and AWS Rekognition provide different models that are used during the analysis. 
The models and their details are shown and summarized in the Supplementary Material, Table 1, Appendix B. 
The selected services also provide some suggestions, called \textit{Rules of reference}, for the correct use of models: they are summarized and reported in the Supplementary Material, Appendix A. 
The tests were operationalized using the SDKs for Python provided by AWSR and ClarifAI\footnote{\url{https://docs.aws.amazon.com/rekognition/latest/dg/labels-detect-labels-image.html}} \footnote{\url{https://web.archive.org/web/20211130220210/https://docs.clarifai.com/api-guide/predict/images}}. 
The output of the models is a JSON file containing different types of information.

According to the research questions codified in Section \ref{sec:rq}, this paper analyses three different tasks: gender recognition, age prediction and image recognition.
As previously mentioned, gender is categorized in a binary format: male and female.
Age prediction is performed differently by the two models.
ClarifAI, assigns a probability to each of the possible age intervals for each image\footnote{Table 2 of the Appendix B in the Supplementary Material.}: the predicted age interval is the one with the highest probability.
Instead, AWSR predicts the age by assigning to each image a specific range from a \textit{'Low'} value to a \textit{'High'} value.
The consequence of such an output format is that it is not possible to obtain the same specific ranges for both models.
According to the rule of reference number 1 of the AWSR\footnote{Appendix A.1 in the Supplementary Material}, the mathematical mean of the predicted range is taken as the final output value of age.
Image recognition models predict concepts, labels, themes and image properties. 
Analysis of this task provides insight into model training, in particular how the images in the training set were labelled. 
ClarifAI provides a model (\textit{general-image-recognition}), that outputs 20 different \textit{concepts} for each image.
Each \textit{concept} has a corresponding probability value. 
The AWSR model (\textit{detect labels}) assigns different labels to each image.
We will refer to the \textit{concepts} of ClarifAI as "labels" for the results of both models, ClarifAI and AWSR.
A maximum of 20 labels are predicted for each image.
For ClarifAI, the labels were grouped into ad-hoc categories. 
Instead, the AWSR model does the grouping itself by specifying pre-defined categories (as mentioned in its documentation).

All ClarifAI output labels were reviewed, analysed and then grouped according to their meaning and main theme.
In the end, the following categories were created \textit{aesthetic}, \textit{education}, \textit{person description}.
The labels of the first two categories are adjectives related to the people depicted in the photos.
\textit{Person Description} contains the same labels as the homonymous AWSR category.
The similar categories of the AWSR model (\textit{Clothing and Accessories}, \textit{Beauty and Personal Care} and \textit{Education}) contain mainly names of objects and descriptors of the images, which are not considered in the current analysis.

The peculiarity of the AWSR output labels is that they are mostly descriptive.
The list of labels is similar to a list of objects that the algorithm recognizes in the image. 
For example, considering one of its categories, \textit{Apparel and Accessories}, some labels are: Jeans, T-shirt, Hat, Shoes etc. 
Instead, looking at the predicted labels from the ClarifAI model, the labels are mainly adjectives.
Adjectives can be much more ethically dangerous than nouns. 
Therefore, we focused our analysis on the \textit{Aesthetics}, \textit{Education}, \textit{Person descriptors} categories of ClarifAI and on the category \textit{Person descriptors} of the AWSR model.

\section{Results and Discussion}\label{sec:results}

\subsection{RQ1.a - Gender Recognition}\label{sec:gender_recognition}

Table \ref{tab:gender-results} presents the values of \textit{Accuracy}, \textit{Recall}, \textit{Precision} and \textit{F1-score}, for both gender recognition models and groups.
The accuracy scores of the EG were lower than those of the CG for both models: the discrepancy between these scores was about 7\% and 4\% respectively.
In general, the results of the EG were lower than those of the CG for both models. 
The \textit{F1-scores} of the EG were lower than those of the CG.
Female \textit{Precision} and male \textit{Recall} showed lower values between EG and CG.

\begin{figure}[t]
    \centering
    \begin{subfigure}[b]{\linewidth}
        \includegraphics[width=\linewidth]{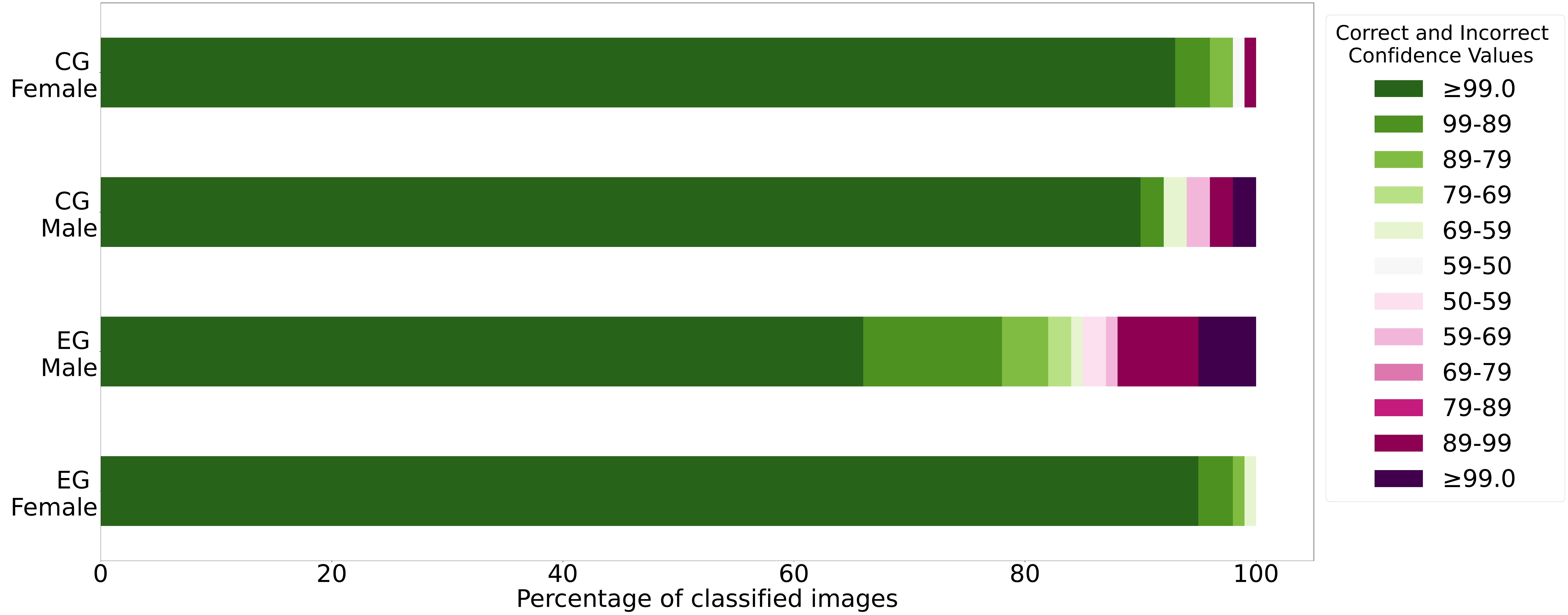}
        \caption{
            Confidence values computed by the AWSR model.
        }
        \label{fig:AWS_conf}       
    \end{subfigure}
    \\
    \begin{subfigure}[b]{\linewidth}
        \includegraphics[width=\linewidth]{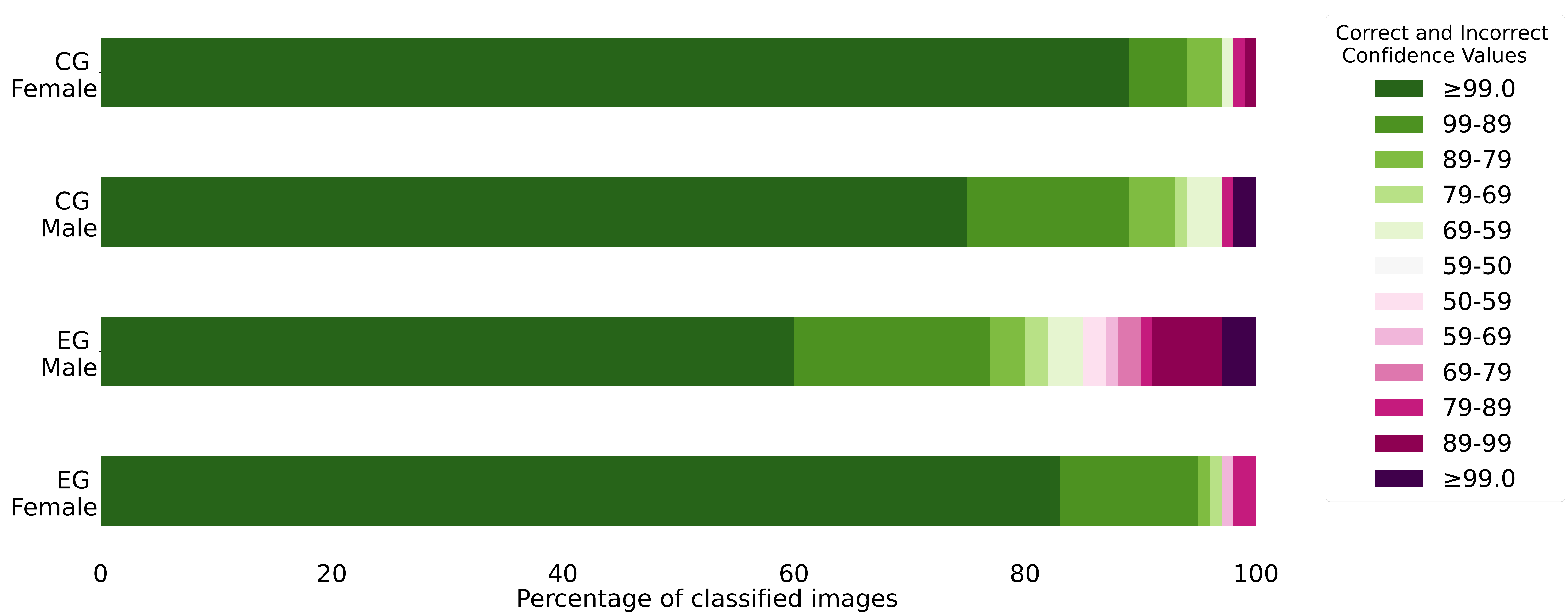}
        \caption{
            Confidence values computed by the ClarifAI model.
        }
        \label{fig:ClarifAI_conf}
    \end{subfigure}
    \caption{Confidence values for each category, computed by the AWSR and ClarifAI models, regarding the gender prediction task. Green bars refer to correct prediction, purple bars refer to incorrect predictions.}
    \label{fig:confidence_values}
\end{figure}

A closer examination of the misclassified images was carried out.
On the one hand, all misclassified images of the EG male group represent children and adolescents. 
On the other hand, the misclassified images of EG females by the ClarifAI model represent old people.
The rule of reference number 2 (Appendix A.1 in the Supplementary Material) regarding the AWSR model suggests that the confidence value assigned to each prediction of gender should be checked and taken into account.
The threshold considered safe for sensible subjects is set at 99.00\% by the rules of the model.
Following the previous recommendation, a detailed examination of the confidence values is carried out on each group of the dataset, as shown in Figure \ref{fig:confidence_values}.

\begin{table}[t]
\centering
\caption{Gender recognition results.}
\label{tab:gender-results}
\begin{tabular}{ll|cccc|cccc|}
\cline{3-10}
                                                     &                 & \multicolumn{4}{c|}{\textbf{Experimental group}}                                                                                                               & \multicolumn{4}{c|}{\textbf{Control group}}                                                                                                                   \\ \hline
\multicolumn{1}{|l|}{\textbf{Model}}                 & \textbf{Gender} & \multicolumn{1}{l|}{\textbf{Acc.}}         & \multicolumn{1}{l|}{\textbf{Prec.}} & \multicolumn{1}{l|}{\textbf{Recall}} & \multicolumn{1}{l|}{\textbf{F1-score}} & \multicolumn{1}{l|}{\textbf{Acc.}}         & \multicolumn{1}{l|}{\textbf{Prec.}} & \multicolumn{1}{l|}{\textbf{Recall}} & \multicolumn{1}{l|}{\textbf{F1-score}} \\ \hline
\multicolumn{1}{|l|}{\multirow{2}{*}{\textbf{AWSR}}} & \textit{Female} & \multicolumn{1}{c|}{\multirow{2}{*}{93\%}} & \multicolumn{1}{c|}{87\%}           & \multicolumn{1}{c|}{100\%}           & 93\%                                 & \multicolumn{1}{c|}{\multirow{2}{*}{97\%}} & \multicolumn{1}{c|}{94\%}          & \multicolumn{1}{c|}{99\%}            & 96\%                                 \\ \cline{2-2} \cline{4-6} \cline{8-10} 
\multicolumn{1}{|l|}{}                               & \textit{Male}   & \multicolumn{1}{c|}{}                      & \multicolumn{1}{c|}{100\%}          & \multicolumn{1}{c|}{85\%}            & 92\%                                 & \multicolumn{1}{c|}{}                      & \multicolumn{1}{c|}{99\%}          & \multicolumn{1}{c|}{94\%}            & 96\%                                 \\ \hline
\multicolumn{1}{|l|}{\multirow{2}{*}{\textbf{CLAI}}} & \textit{Female} & \multicolumn{1}{c|}{\multirow{2}{*}{91\%}} & \multicolumn{1}{c|}{87\%}           & \multicolumn{1}{c|}{97\%}            & 92\%                                 & \multicolumn{1}{c|}{\multirow{2}{*}{98\%}} & \multicolumn{1}{c|}{97\%}          & \multicolumn{1}{c|}{98\%}            & 98\%                                 \\ \cline{2-2} \cline{4-6} \cline{8-10} 
\multicolumn{1}{|l|}{}                               & \textit{Male}   & \multicolumn{1}{c|}{}                      & \multicolumn{1}{c|}{97\%}           & \multicolumn{1}{c|}{85\%}            & 90\%                                 & \multicolumn{1}{c|}{}                      & \multicolumn{1}{c|}{98\%}          & \multicolumn{1}{c|}{97\%}            & 97\%                                 \\ \hline
\end{tabular}
\end{table}

Figures \ref{fig:AWS_conf} and \ref{fig:ClarifAI_conf} illustrate the distribution of confidence values for each class of the dataset, including the values of correct and incorrect predictions: the colours green and purple represent the correctness and incorrectness of the model prediction in the study, respectively.
The different shades of these colours represent the confidence levels and their own level of error.
Both models performed poorly for the EG male category.
The AWSR model correctly classified 66\% of the images with a confidence level greater than or equal to 99.00\%. 
The ClarifAI model correctly classified 60\% of the images with a confidence level greater than or equal to 99.00\%. 
We observe that the categories with the best performance were the female ones.
In particular the category EG female, for the AWSR model, does not contain misclassification of gender. 
Looking at the incorrect predictions, we found that  about a 5\% of the predictions of the EG male classes had a confidence value greater than or equal to 99.00\%. 
This means that the model misclassified images for which it was very confident about the prediction.

Finally, Table \ref{tab:cf-results} shows the accuracy values considering only the prediction with a confidence value greater or equal to 99.00\%.
Within females, the discrepancy between EG and CG was reduced.
In fact, in the case of ClarifAI, the difference between EG and CG is about 6\%, while in the case of AWSR, EG performed better than CG (2\%).
The males highlight large performance discrepancies between EG and CG. 
In the AWSR case, the difference in accuracy was 24\%, while in the case of ClarifAI the difference in accuracy was equal to 15\%.
\vspace{0.5cm}

\framebox{
    \begin{minipage}[h]{0.90\linewidth}
        We observed that the prediction of gender performed differently between EG and CG.
        The results showed a high error rate towards the EG male category.
        In particular, both models correctly predicted 85\% of the images in the EG male class, in contrast to 97\% and 94\% of the CG male class for ClarifAI and AWSR respectively.
    \end{minipage}
}

\begin{table}[t]
    \centering
    \caption{
        Accuracy values for the gender prediction task, considering only the prediction with a confidence value greater than or equal to 99.00\%. 
    }
    \label{tab:cf-results}
    \resizebox{\columnwidth}{!}{%
\begin{tabular}{l|cc|cc|}
\cline{2-5}
\multirow{2}{*}{} & \multicolumn{2}{c|}{\textbf{AWSR}} & \multicolumn{2}{c|}{\textbf{ClarifAI}}                       \\ \cline{2-5} 
& \multicolumn{1}{l|}{Experimental group} & \multicolumn{1}{l|}{Control group} & \multicolumn{1}{l|}{Experimental group} & \multicolumn{1}{l|}{Control group} \\ \hline
\multicolumn{1}{|l|}{\textbf{Female}} & \multicolumn{1}{c|}{95\%}                   & 93\%                                     & \multicolumn{1}{c|}{83\%}                   & 89\%                                     \\ \hline
\multicolumn{1}{|l|}{\textbf{Male}}   & \multicolumn{1}{c|}{66\%}                   & 90\%                                     & \multicolumn{1}{c|}{60\%}                   & 75\%                                     \\ \hline
\end{tabular}%
}
\end{table}

\subsection{RQ1.b - Age Prediction}\label{sec:age_prediction}
In the models analysed, age prediction is a classification problem and the result of the prediction consists of a range of ages rather than a precise value, as described in section \ref{sec:models}. 
\begin{table}[t]
\footnotesize
    \centering
    \caption{
        Accuracy values for the age prediction task. 
    }
    \label{tab:age_total_results}
    \resizebox{\columnwidth}{!}{%
\begin{tabular}{l|cc|cc|}
\cline{2-5}
\multicolumn{1}{c|}{}                        & \multicolumn{2}{c|}{\textbf{AWSR}}                                 & \multicolumn{2}{c|}{\textbf{ClarifAI}}                            \\ \cline{2-5} 
\multicolumn{1}{c|}{}                        & \multicolumn{1}{c|}{Experimental group} & Control group & \multicolumn{1}{c|}{Experimental group} & Control group \\ \hline
\multicolumn{1}{|l|}{\textbf{Correct Age}}   & \multicolumn{1}{c|}{52\%}                & 45\%                   & \multicolumn{1}{c|}{48\%}                & 49\%                   \\ \hline
\multicolumn{1}{|l|}{\textbf{Incorrect Age}} & \multicolumn{1}{c|}{48\%}                & 55\%                   & \multicolumn{1}{c|}{52\%}                & 51\%                   \\ \hline
\end{tabular}%
}
\end{table}
The accuracy values are presented in Table \ref{tab:age_total_results}.
The results show that the performances of both models are low: only half of the samples are predicted correctly. 

The truth tables were constructed using the ClarifAI ranges for both the true range and the predicted range.
Each value of the truth tables \ref{tab:truth_table_eg_clarifai}, \ref{tab:truth_table_cg_clarifai}, \ref{tab:truth_table_eg_aws}, \ref{tab:truth_table_cg_aws} represent the number of images predicted in the corresponding range of ages.
Looking at the ClarifAI model (Tables \ref{tab:truth_table_cg_clarifai} and \ref{tab:truth_table_eg_clarifai}) we can observe some differences between EG and CG. 
The EG performed worst in the ranges: 20-29, 30-39, 40-49, while the CG performed worst in the ranges between: 40-49, 50-59, 60-69, $\geq70$.
The AWSR predictions (Tables \ref{tab:truth_table_cg_aws} and \ref{tab:truth_table_eg_aws}) were slightly more accurate.

\begin{table}[t]
    \caption{Truth tables regarding age prediction.}
    \label{tab:truth_table_clarifai}
    \begin{subtable}[ht]{0.49\linewidth}
        \caption{ClarifAI Experimental group.}
        \label{tab:truth_table_eg_clarifai}
        \resizebox{\columnwidth}{!}{%
            
            \begin{tabular}{|lccccccccc|}
            \hline
            \multicolumn{10}{|l|}{\textit{\textbf{Experimental group}}} \\ \hline
            \multicolumn{1}{|c|}{\textbf{}} &
              \multicolumn{9}{c|}{\textbf{PREDICTED RANGE}} \\ \hline
            \multicolumn{1}{|l|}{\textbf{\begin{tabular}[c]{@{}l@{}}TRUE\\ RANGE\end{tabular}}} &
              \multicolumn{1}{l|}{\textbf{0-2}} &
              \multicolumn{1}{l|}{\textbf{3-9}} &
              \multicolumn{1}{l|}{\textbf{10-19}} &
              \multicolumn{1}{l|}{\textbf{20-29}} &
              \multicolumn{1}{l|}{\textbf{30-39}} &
              \multicolumn{1}{l|}{\textbf{40-49}} &
              \multicolumn{1}{l|}{\textbf{50-59}} &
              \multicolumn{1}{l|}{\textbf{60-69}} &
              \multicolumn{1}{l|}{\textbf{\textgreater{}70}} \\ \hline
            \multicolumn{1}{|l|}{\textbf{0 - 2}} &
              \multicolumn{1}{c|}{\cellcolor[HTML]{D9D7D7}2} &
              \multicolumn{1}{c|}{} &
              \multicolumn{1}{c|}{} &
              \multicolumn{1}{c|}{} &
              \multicolumn{1}{c|}{} &
              \multicolumn{1}{c|}{} &
              \multicolumn{1}{c|}{} &
              \multicolumn{1}{c|}{} &
               \\ \hline
            \multicolumn{1}{|l|}{\textbf{3 - 9}} &
              \multicolumn{1}{c|}{3} &
              \multicolumn{1}{c|}{\cellcolor[HTML]{D9D7D7}9} &
              \multicolumn{1}{c|}{} &
              \multicolumn{1}{c|}{} &
              \multicolumn{1}{c|}{} &
              \multicolumn{1}{c|}{} &
              \multicolumn{1}{c|}{} &
              \multicolumn{1}{c|}{} &
               \\ \hline
            \multicolumn{1}{|l|}{\textbf{10 - 19}} &
              \multicolumn{1}{c|}{} &
              \multicolumn{1}{c|}{4} &
              \multicolumn{1}{c|}{\cellcolor[HTML]{D9D7D7}15} &
              \multicolumn{1}{c|}{9} &
              \multicolumn{1}{c|}{} &
              \multicolumn{1}{c|}{} &
              \multicolumn{1}{c|}{} &
              \multicolumn{1}{c|}{} &
               \\ \hline
            \multicolumn{1}{|l|}{\textbf{20 - 29}} &
              \multicolumn{1}{c|}{} &
              \multicolumn{1}{c|}{3} &
              \multicolumn{1}{c|}{14} &
              \multicolumn{1}{c|}{\cellcolor[HTML]{D9D7D7}24} &
              \multicolumn{1}{c|}{10} &
              \multicolumn{1}{c|}{} &
              \multicolumn{1}{c|}{} &
              \multicolumn{1}{c|}{} &
               \\ \hline
            \multicolumn{1}{|l|}{\textbf{30 - 39}} &
              \multicolumn{1}{c|}{} &
              \multicolumn{1}{c|}{2} &
              \multicolumn{1}{c|}{2} &
              \multicolumn{1}{c|}{9} &
              \multicolumn{1}{c|}{\cellcolor[HTML]{D9D7D7}6} &
              \multicolumn{1}{c|}{3} &
              \multicolumn{1}{c|}{} &
              \multicolumn{1}{c|}{} &
               \\ \hline
            \multicolumn{1}{|l|}{\textbf{40 - 49}} &
              \multicolumn{1}{c|}{} &
              \multicolumn{1}{c|}{} &
              \multicolumn{1}{c|}{1} &
              \multicolumn{1}{c|}{2} &
              \multicolumn{1}{c|}{2} &
              \multicolumn{1}{c|}{\cellcolor[HTML]{D9D7D7}6} &
              \multicolumn{1}{c|}{2} &
              \multicolumn{1}{c|}{} &
               \\ \hline
            \multicolumn{1}{|l|}{\textbf{50 - 59}} &
              \multicolumn{1}{c|}{} &
              \multicolumn{1}{c|}{} &
              \multicolumn{1}{c|}{} &
              \multicolumn{1}{c|}{} &
              \multicolumn{1}{c|}{} &
              \multicolumn{1}{c|}{1} &
              \multicolumn{1}{c|}{\cellcolor[HTML]{D9D7D7}0} &
              \multicolumn{1}{c|}{1} &
               \\ \hline
            \multicolumn{1}{|l|}{\textbf{60 - 69}} &
              \multicolumn{1}{l|}{} &
              \multicolumn{1}{l|}{} &
              \multicolumn{1}{l|}{} &
              \multicolumn{1}{l|}{} &
              \multicolumn{1}{l|}{} &
              \multicolumn{1}{l|}{} &
              \multicolumn{1}{l|}{} &
              \multicolumn{1}{l|}{} &
              \multicolumn{1}{l|}{} \\ \hline
            \multicolumn{1}{|l|}{\textbf{\textgreater 70}} &
              \multicolumn{1}{l|}{} &
              \multicolumn{1}{l|}{} &
              \multicolumn{1}{l|}{} &
              \multicolumn{1}{l|}{} &
              \multicolumn{1}{l|}{} &
              \multicolumn{1}{l|}{} &
              \multicolumn{1}{l|}{} &
              \multicolumn{1}{l|}{} &
              \multicolumn{1}{l|}{} \\ \hline
            \end{tabular}%
        }
    \end{subtable}
    \hfill
    \begin{subtable}[ht]{0.49\linewidth}
    \caption{ClarifAI Control group.}
    \label{tab:truth_table_cg_clarifai}
        \resizebox{\columnwidth}{!}{%
            \begin{tabular}{|lccccccccl|}
            \hline
            \multicolumn{10}{|l|}{\textit{\textbf{Control group}}} \\ \hline
            \multicolumn{1}{|l|}{} &
              \multicolumn{9}{c|}{\textbf{PREDICTED RANGE}} \\ \hline
            \multicolumn{1}{|l|}{\textbf{\begin{tabular}[c]{@{}l@{}}TRUE \\ RANGE\end{tabular}}} &
              \multicolumn{1}{l|}{\textbf{0-2}} &
              \multicolumn{1}{l|}{\textbf{3-9}} &
              \multicolumn{1}{l|}{\textbf{10-19}} &
              \multicolumn{1}{l|}{\textbf{20-29}} &
              \multicolumn{1}{l|}{\textbf{30-39}} &
              \multicolumn{1}{l|}{\textbf{40-49}} &
              \multicolumn{1}{l|}{\textbf{50-59}} &
              \multicolumn{1}{l|}{\textbf{60-69}} &
              \textbf{\textgreater{}70} \\ \hline
            \multicolumn{1}{|l|}{\textbf{0 - 2}} &
              \multicolumn{1}{c|}{\cellcolor[HTML]{D9D7D7}0} &
              \multicolumn{1}{l|}{} &
              \multicolumn{1}{l|}{} &
              \multicolumn{1}{l|}{} &
              \multicolumn{1}{l|}{} &
              \multicolumn{1}{l|}{} &
              \multicolumn{1}{l|}{} &
              \multicolumn{1}{l|}{} &
               \\ \hline
            \multicolumn{1}{|l|}{\textbf{3 - 9}} &
              \multicolumn{1}{l|}{} &
              \multicolumn{1}{c|}{\cellcolor[HTML]{D9D7D7}11} &
              \multicolumn{1}{c|}{4} &
              \multicolumn{1}{l|}{} &
              \multicolumn{1}{l|}{} &
              \multicolumn{1}{l|}{} &
              \multicolumn{1}{l|}{} &
              \multicolumn{1}{l|}{} &
               \\ \hline
            \multicolumn{1}{|l|}{\textbf{10 - 19}} &
              \multicolumn{1}{l|}{} &
              \multicolumn{1}{c|}{5} &
              \multicolumn{1}{c|}{\cellcolor[HTML]{D9D7D7}13} &
              \multicolumn{1}{c|}{7} &
              \multicolumn{1}{l|}{} &
              \multicolumn{1}{l|}{} &
              \multicolumn{1}{l|}{} &
              \multicolumn{1}{l|}{} &
               \\ \hline
            \multicolumn{1}{|l|}{\textbf{20 - 29}} &
              \multicolumn{1}{l|}{} &
              \multicolumn{1}{l|}{} &
              \multicolumn{1}{c|}{2} &
              \multicolumn{1}{c|}{\cellcolor[HTML]{D9D7D7}20} &
              \multicolumn{1}{c|}{3} &
              \multicolumn{1}{l|}{} &
              \multicolumn{1}{l|}{} &
              \multicolumn{1}{l|}{} &
               \\ \hline
            \multicolumn{1}{|l|}{\textbf{30 - 39}} &
              \multicolumn{1}{l|}{} &
              \multicolumn{1}{l|}{} &
              \multicolumn{1}{l|}{} &
              \multicolumn{1}{c|}{9} &
              \multicolumn{1}{c|}{\cellcolor[HTML]{D9D7D7}15} &
              \multicolumn{1}{l|}{} &
              \multicolumn{1}{l|}{} &
              \multicolumn{1}{l|}{} &
               \\ \hline
            \multicolumn{1}{|l|}{\textbf{40 - 49}} &
              \multicolumn{1}{l|}{} &
              \multicolumn{1}{l|}{} &
              \multicolumn{1}{l|}{} &
              \multicolumn{1}{c|}{4} &
              \multicolumn{1}{c|}{9} &
              \multicolumn{1}{c|}{\cellcolor[HTML]{D9D7D7}11} &
              \multicolumn{1}{c|}{3} &
              \multicolumn{1}{l|}{} &
               \\ \hline
            \multicolumn{1}{|l|}{\textbf{50 - 59}} &
              \multicolumn{1}{l|}{} &
              \multicolumn{1}{l|}{} &
              \multicolumn{1}{l|}{} &
              \multicolumn{1}{c|}{3} &
              \multicolumn{1}{c|}{6} &
              \multicolumn{1}{c|}{9} &
              \multicolumn{1}{c|}{\cellcolor[HTML]{D9D7D7}10} &
              \multicolumn{1}{c|}{3} &
               \\ \hline
            \multicolumn{1}{|l|}{\textbf{60 - 69}} &
              \multicolumn{1}{l|}{} &
              \multicolumn{1}{l|}{} &
              \multicolumn{1}{l|}{} &
              \multicolumn{1}{l|}{} &
              \multicolumn{1}{c|}{3} &
              \multicolumn{1}{c|}{4} &
              \multicolumn{1}{c|}{12} &
              \multicolumn{1}{c|}{\cellcolor[HTML]{D9D7D7}9} &
               \\ \hline
            \multicolumn{1}{|l|}{\textbf{\textgreater 70}} &
              \multicolumn{1}{l|}{} &
              \multicolumn{1}{l|}{} &
              \multicolumn{1}{l|}{} &
              \multicolumn{1}{c|}{1} &
              \multicolumn{1}{l|}{} &
              \multicolumn{1}{l|}{} &
              \multicolumn{1}{c|}{7} &
              \multicolumn{1}{c|}{11} &
              \cellcolor[HTML]{D9D7D7}6 \\ \hline
            \end{tabular}%
        }
    \end{subtable}
    \\
    \begin{subtable}[ht]{0.49\linewidth}
        \caption{AWSR experimental group.}
        \label{tab:truth_table_eg_aws}
        \resizebox{\columnwidth}{!}{%
            \begin{tabular}{|lccccccccc|}
            \hline
            \multicolumn{10}{|l|}{\textit{\textbf{Experimental group}}} \\ \hline
            \multicolumn{1}{|l|}{} &
              \multicolumn{9}{c|}{\textbf{PREDICTED RANGE}} \\ \hline
            \multicolumn{1}{|l|}{\textbf{\begin{tabular}[c]{@{}l@{}}TRUE \\ RANGE\end{tabular}}} &
              \multicolumn{1}{l|}{\textbf{0-2}} &
              \multicolumn{1}{l|}{\textbf{3-9}} &
              \multicolumn{1}{l|}{\textbf{10-19}} &
              \multicolumn{1}{l|}{\textbf{20-29}} &
              \multicolumn{1}{l|}{\textbf{30-39}} &
              \multicolumn{1}{l|}{\textbf{40-49}} &
              \multicolumn{1}{l|}{\textbf{50-59}} &
              \multicolumn{1}{l|}{\textbf{60-69}} &
              \textbf{\textgreater{}70} \\ \hline
            \multicolumn{1}{|l|}{\textbf{0 - 2}} &
              \multicolumn{1}{c|}{\cellcolor[HTML]{D9D7D7}1} &
              \multicolumn{1}{l|}{1} &
              \multicolumn{1}{l|}{} &
              \multicolumn{1}{l|}{} &
              \multicolumn{1}{l|}{} &
              \multicolumn{1}{l|}{} &
              \multicolumn{1}{l|}{} &
              \multicolumn{1}{l|}{} &
               \\ \hline
            \multicolumn{1}{|l|}{\textbf{3 - 9}} &
              \multicolumn{1}{c|}{1} &
              \multicolumn{1}{c|}{\cellcolor[HTML]{D9D7D7}10} &
              \multicolumn{1}{c|}{1} &
              \multicolumn{1}{l|}{} &
              \multicolumn{1}{l|}{} &
              \multicolumn{1}{l|}{} &
              \multicolumn{1}{l|}{} &
              \multicolumn{1}{l|}{} &
               \\ \hline
            \multicolumn{1}{|l|}{\textbf{10 - 19}} &
              \multicolumn{1}{c|}{} &
              \multicolumn{1}{c|}{3} &
              \multicolumn{1}{c|}{\cellcolor[HTML]{D9D7D7}14} &
              \multicolumn{1}{c|}{9} &
              \multicolumn{1}{l|}{} &
              \multicolumn{1}{l|}{} &
              \multicolumn{1}{l|}{} &
              \multicolumn{1}{l|}{} &
               \\ \hline
            \multicolumn{1}{|l|}{\textbf{20 - 29}} &
              \multicolumn{1}{l|}{} &
              \multicolumn{1}{l|}{} &
              \multicolumn{1}{c|}{15} &
              \multicolumn{1}{c|}{\cellcolor[HTML]{D9D7D7}22} &
              \multicolumn{1}{c|}{12} &
              \multicolumn{1}{c|}{2} &
              \multicolumn{1}{l|}{} &
              \multicolumn{1}{l|}{} &
               \\ \hline
            \multicolumn{1}{|l|}{\textbf{30 - 39}} &
              \multicolumn{1}{l|}{} &
              \multicolumn{1}{c|}{1} &
              \multicolumn{1}{c|}{2} &
              \multicolumn{1}{c|}{10} &
              \multicolumn{1}{c|}{\cellcolor[HTML]{D9D7D7}9} &
              \multicolumn{1}{l|}{} &
              \multicolumn{1}{l|}{} &
              \multicolumn{1}{l|}{} &
               \\ \hline
            \multicolumn{1}{|l|}{\textbf{40 - 49}} &
              \multicolumn{1}{l|}{} &
              \multicolumn{1}{l|}{} &
              \multicolumn{1}{l|}{} &
              \multicolumn{1}{c|}{2} &
              \multicolumn{1}{c|}{5} &
              \multicolumn{1}{c|}{\cellcolor[HTML]{D9D7D7}4} &
              \multicolumn{1}{c|}{2} &
              \multicolumn{1}{l|}{} &
               \\ \hline
            \multicolumn{1}{|l|}{\textbf{50 - 59}} &
              \multicolumn{1}{l|}{} &
              \multicolumn{1}{l|}{} &
              \multicolumn{1}{l|}{} &
              \multicolumn{1}{l|}{} &
              \multicolumn{1}{l|}{} &
              \multicolumn{1}{c|}{1} &
              \multicolumn{1}{c|}{\cellcolor[HTML]{D9D7D7}1} &
              \multicolumn{1}{l|}{} &
               \\ \hline
            \multicolumn{1}{|l|}{\textbf{60 - 69}} &
              \multicolumn{1}{l|}{} &
              \multicolumn{1}{l|}{} &
              \multicolumn{1}{l|}{} &
              \multicolumn{1}{l|}{} &
              \multicolumn{1}{l|}{} &
              \multicolumn{1}{l|}{} &
              \multicolumn{1}{l|}{} &
              \multicolumn{1}{l|}{} &
               \\ \hline
            \multicolumn{1}{|l|}{\textbf{\textgreater 70}} &
              \multicolumn{1}{l|}{} &
              \multicolumn{1}{l|}{} &
              \multicolumn{1}{l|}{} &
              \multicolumn{1}{l|}{} &
              \multicolumn{1}{l|}{} &
              \multicolumn{1}{l|}{} &
              \multicolumn{1}{l|}{} &
              \multicolumn{1}{l|}{} &
               \\ \hline
            \end{tabular}%
        }
    \end{subtable}
    \hfil
    \begin{subtable}[ht]{0.49\linewidth}
        \caption{AWSR control group.}
        \label{tab:truth_table_cg_aws}
        \resizebox{\columnwidth}{!}{%
            \begin{tabular}{|lccccccccc|}
            \hline
            \multicolumn{10}{|l|}{\textit{\textbf{Control group}}} \\ \hline
            \multicolumn{1}{|l|}{} &
              \multicolumn{9}{c|}{\textbf{PREDICTED RANGE}} \\ \hline
            \multicolumn{1}{|l|}{\textbf{\begin{tabular}[c]{@{}l@{}}TRUE \\ RANGE\end{tabular}}} &
              \multicolumn{1}{l|}{\textbf{0-2}} &
              \multicolumn{1}{l|}{\textbf{3-9}} &
              \multicolumn{1}{l|}{\textbf{10-19}} &
              \multicolumn{1}{l|}{\textbf{20-29}} &
              \multicolumn{1}{l|}{\textbf{30-39}} &
              \multicolumn{1}{l|}{\textbf{40-49}} &
              \multicolumn{1}{l|}{\textbf{50-59}} &
              \multicolumn{1}{l|}{\textbf{60-69}} &
              \multicolumn{1}{l|}{\textbf{\textgreater{}70}} \\ \hline
            \multicolumn{1}{|l|}{\textbf{0 - 2}} &
              \multicolumn{1}{c|}{\cellcolor[HTML]{D9D7D7}0} &
              \multicolumn{1}{c|}{} &
              \multicolumn{1}{c|}{} &
              \multicolumn{1}{c|}{} &
              \multicolumn{1}{c|}{} &
              \multicolumn{1}{c|}{} &
              \multicolumn{1}{c|}{} &
              \multicolumn{1}{c|}{} &
               \\ \hline
            \multicolumn{1}{|l|}{\textbf{3 - 9}} &
              \multicolumn{1}{c|}{1} &
              \multicolumn{1}{c|}{\cellcolor[HTML]{D9D7D7}9} &
              \multicolumn{1}{c|}{5} &
              \multicolumn{1}{c|}{} &
              \multicolumn{1}{c|}{} &
              \multicolumn{1}{c|}{} &
              \multicolumn{1}{c|}{} &
              \multicolumn{1}{c|}{} &
               \\ \hline
            \multicolumn{1}{|l|}{\textbf{10 - 19}} &
              \multicolumn{1}{c|}{} &
              \multicolumn{1}{c|}{3} &
              \multicolumn{1}{c|}{\cellcolor[HTML]{D9D7D7}12} &
              \multicolumn{1}{c|}{10} &
              \multicolumn{1}{c|}{} &
              \multicolumn{1}{c|}{} &
              \multicolumn{1}{c|}{} &
              \multicolumn{1}{c|}{} &
               \\ \hline
            \multicolumn{1}{|l|}{\textbf{20 - 29}} &
              \multicolumn{1}{c|}{} &
              \multicolumn{1}{c|}{} &
              \multicolumn{1}{c|}{3} &
              \multicolumn{1}{c|}{\cellcolor[HTML]{D9D7D7}21} &
              \multicolumn{1}{c|}{1} &
              \multicolumn{1}{c|}{} &
              \multicolumn{1}{c|}{} &
              \multicolumn{1}{c|}{} &
               \\ \hline
            \multicolumn{1}{|l|}{\textbf{30 - 39}} &
              \multicolumn{1}{c|}{} &
              \multicolumn{1}{c|}{} &
              \multicolumn{1}{c|}{1} &
              \multicolumn{1}{c|}{10} &
              \multicolumn{1}{c|}{\cellcolor[HTML]{D9D7D7}7} &
              \multicolumn{1}{c|}{3} &
              \multicolumn{1}{c|}{} &
              \multicolumn{1}{c|}{} &
               \\ \hline
            \multicolumn{1}{|l|}{\textbf{40 - 49}} &
              \multicolumn{1}{c|}{} &
              \multicolumn{1}{c|}{} &
              \multicolumn{1}{c|}{} &
              \multicolumn{1}{c|}{2} &
              \multicolumn{1}{c|}{6} &
              \multicolumn{1}{c|}{\cellcolor[HTML]{D9D7D7}17} &
              \multicolumn{1}{c|}{2} &
              \multicolumn{1}{c|}{} &
               \\ \hline
            \multicolumn{1}{|l|}{\textbf{50 - 59}} &
              \multicolumn{1}{c|}{} &
              \multicolumn{1}{c|}{} &
              \multicolumn{1}{c|}{} &
              \multicolumn{1}{c|}{} &
              \multicolumn{1}{c|}{2} &
              \multicolumn{1}{c|}{12} &
              \multicolumn{1}{c|}{\cellcolor[HTML]{D9D7D7}17} &
              \multicolumn{1}{c|}{} &
               \\ \hline
            \multicolumn{1}{|l|}{\textbf{60 - 69}} &
              \multicolumn{1}{c|}{} &
              \multicolumn{1}{c|}{} &
              \multicolumn{1}{c|}{} &
              \multicolumn{1}{c|}{} &
              \multicolumn{1}{c|}{} &
              \multicolumn{1}{c|}{4} &
              \multicolumn{1}{c|}{15} &
              \multicolumn{1}{c|}{\cellcolor[HTML]{D9D7D7}6} &
              1 \\ \hline
            \multicolumn{1}{|l|}{\textbf{\textgreater 70}} &
              \multicolumn{1}{c|}{} &
              \multicolumn{1}{c|}{} &
              \multicolumn{1}{c|}{} &
              \multicolumn{1}{c|}{} &
              \multicolumn{1}{c|}{} &
              \multicolumn{1}{c|}{} &
              \multicolumn{1}{c|}{6} &
              \multicolumn{1}{c|}{15} &
              \cellcolor[HTML]{D9D7D7}3 \\ \hline
            \end{tabular}%
        }
    \end{subtable}
\end{table}

Most of the errors for the EG were in the ranges: 30-39, 40-49 whereas for the CG they were in the ranges: 30-39, 60-69, $\geq70$.
The comparison should take into account that the average life expectancy of people with Down syndrome is around 65 years, and the EG dataset contains images of people with a maximum age in the range 50-59.

Focusing on the predictions of ClarifAI on the EG, Table \ref{tab:truth_table_eg_clarifai} shows that the predicted ranges 3-9 and 10-19 are the two intervals with higher variance in the dataset.
This means that some images with a true age range of 20-29 and 30-39 are labelled with the age range 3-9 or 10-19.
The same can be verified for the age range 40-49 with predictions of 10-19.
Differently, for the CG, Table \ref{tab:truth_table_cg_clarifai}, the age ranges with higher variance are 20-29 and 30-39.
Some images with a true age range of 50-59 or 60-69 are labelled with age ranges of 20-29 and 30-39.
For the CG, the predicted age ranges are lower than the real ones, but they never coincide with the age ranges of children.
The predictions corresponding to a lower age could be due to the fact that the dataset is made up of images of famous people, i.e. people with facial surgery and make-up who make themselves look younger.
Instead, the images representing adult people with Down syndrome are classified with children in age ranges such as 3-9 and 10-19.

The AWSR model is more stable in the ranges of its predictions.
The ranges with higher variance for the EG are 3-9 and 10-19, whereas for the CG almost all ranges have the same variance. However, some errors in the AWSR model are quite similar to those in the ClarifAI model for the experimental group.
Some images with a true age range of 30-39 are classified as having an age range of 3-9 or 10-19.
\vspace{0.5cm}

\framebox{
    \begin{minipage}[h]{0.90\linewidth}
        We observed that there are some differences in age estimation between EG and CG. The results lead to the conclusion that both models assign the age range of children to adults belonging to the EG.
    \end{minipage}
}

\subsection{RQ2. Image labelling}\label{sec:general_image_recognition}

\subsubsection{Aesthetics and Education}\label{sec:general_aesthetics}
\begin{figure}[ht]
    \centering
    \begin{subfigure}[b]{0.49\linewidth}
        \includegraphics[width=\linewidth]{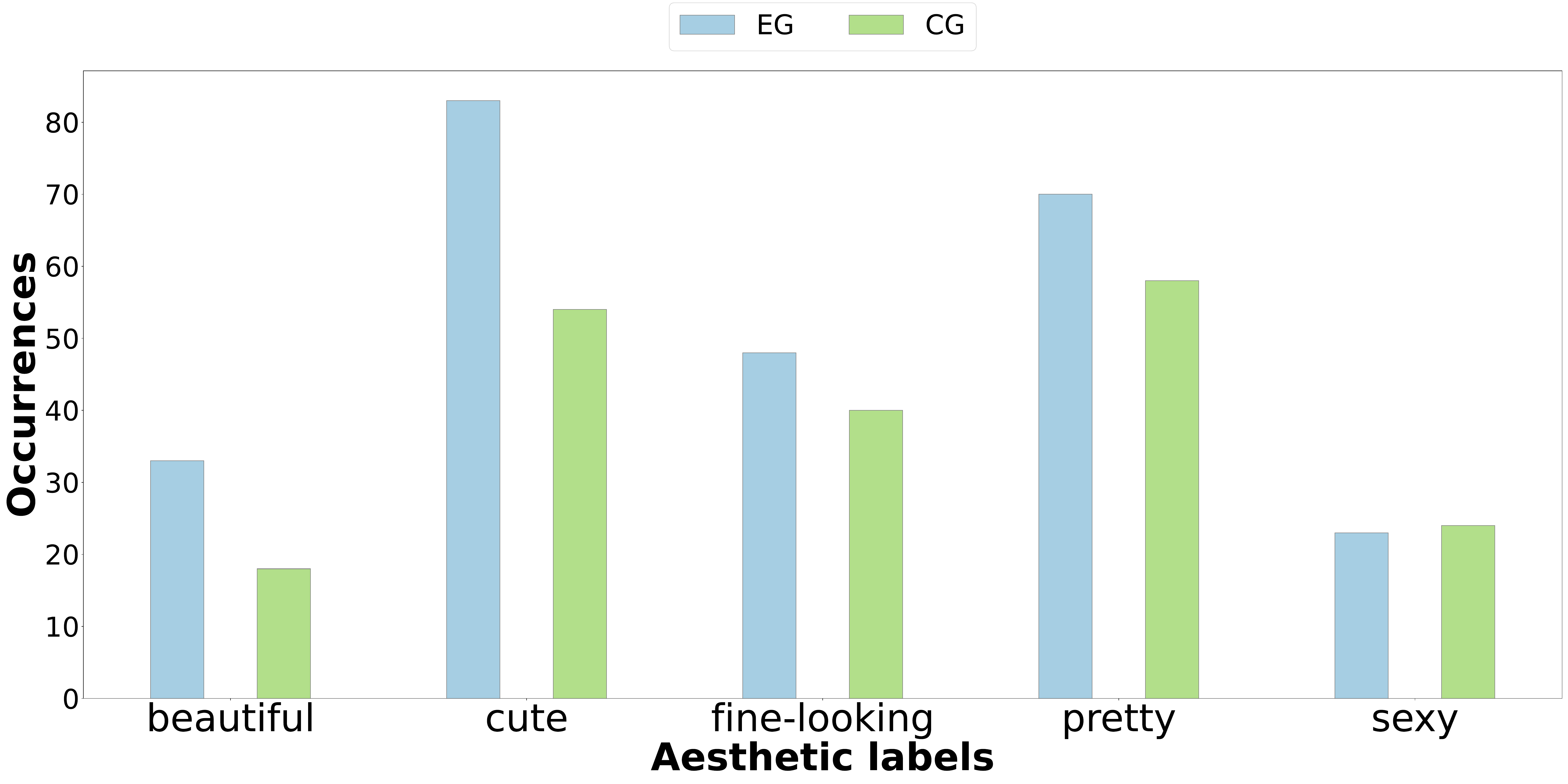}
        \caption{
            Comparison between experimental group and control group.
        }
        \label{fig:aesthetic_Clarifai}
    \end{subfigure}
    \hfill
    \begin{subfigure}[b]{0.49\linewidth}
        \centering
        \includegraphics[width=\linewidth]{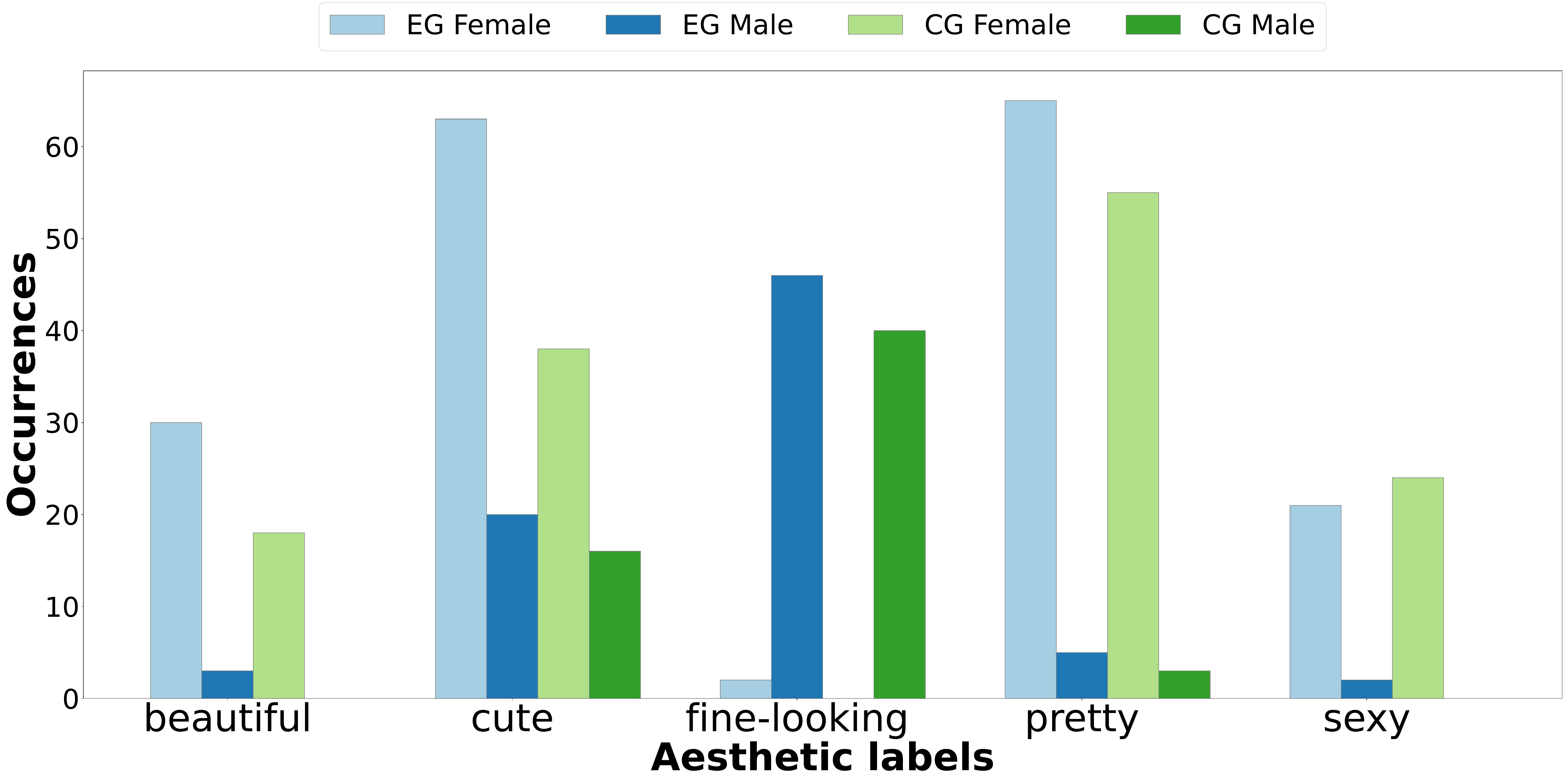}
        \caption{
            Gendered comparison between experimental group and control group.
        }
        \label{fig:aesthetic_Gendered_Clarifai}
    \end{subfigure}
    \caption{Comparison regarding \textit{Aesthetic} labels assigned by the ClarifAI model}
\end{figure}

\begin{figure}[ht]
    \begin{subfigure}[b]{0.48\linewidth}
        \centering
        \includegraphics[width=\linewidth]{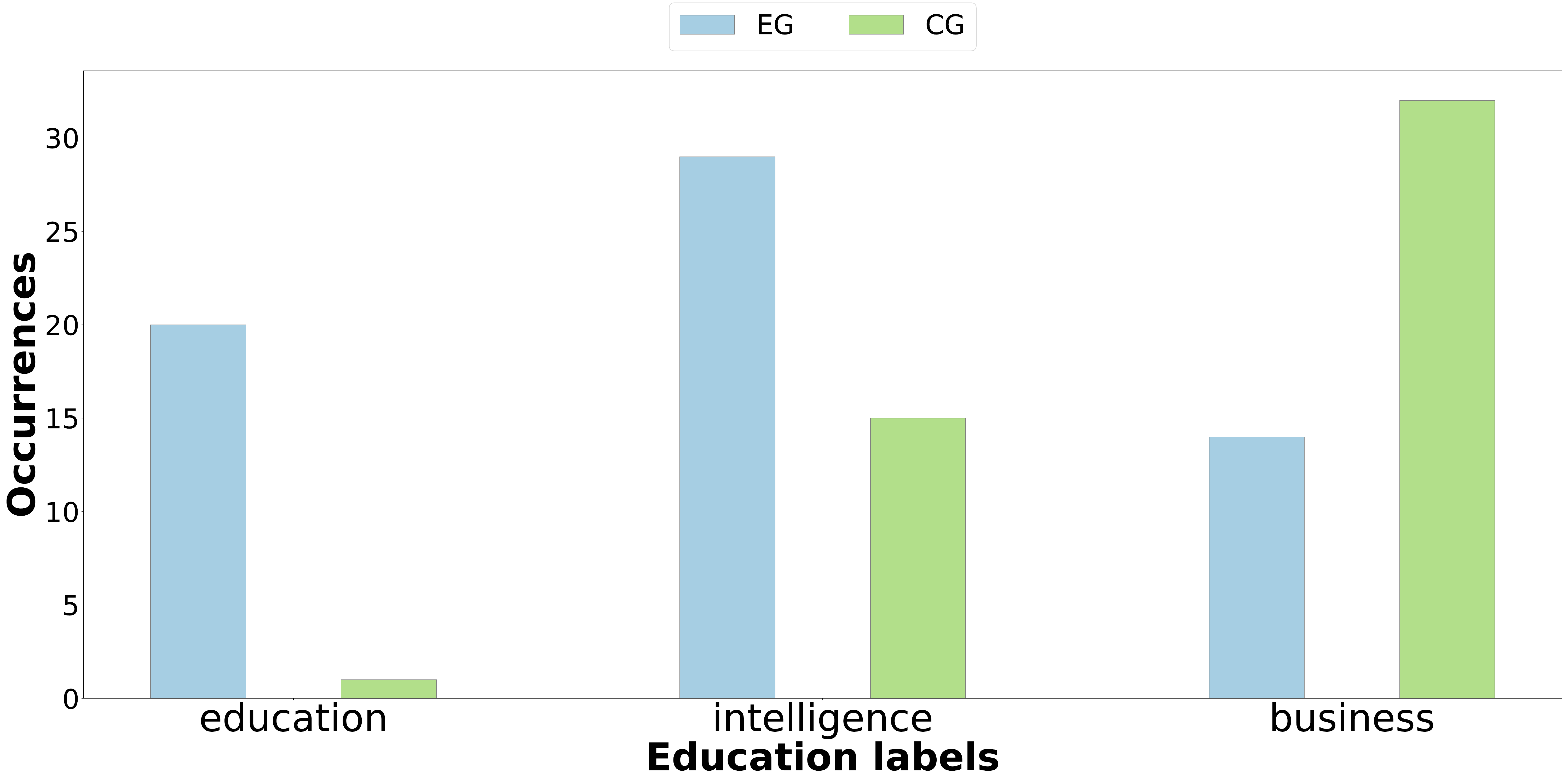}
        \caption{
            Comparison between experimental group and control group regarding.
        }
        \label{fig:education_Clarifai}
    \end{subfigure}
    \hfill
    \begin{subfigure}[b]{0.49\linewidth}
        \centering
        \includegraphics[width=\linewidth]{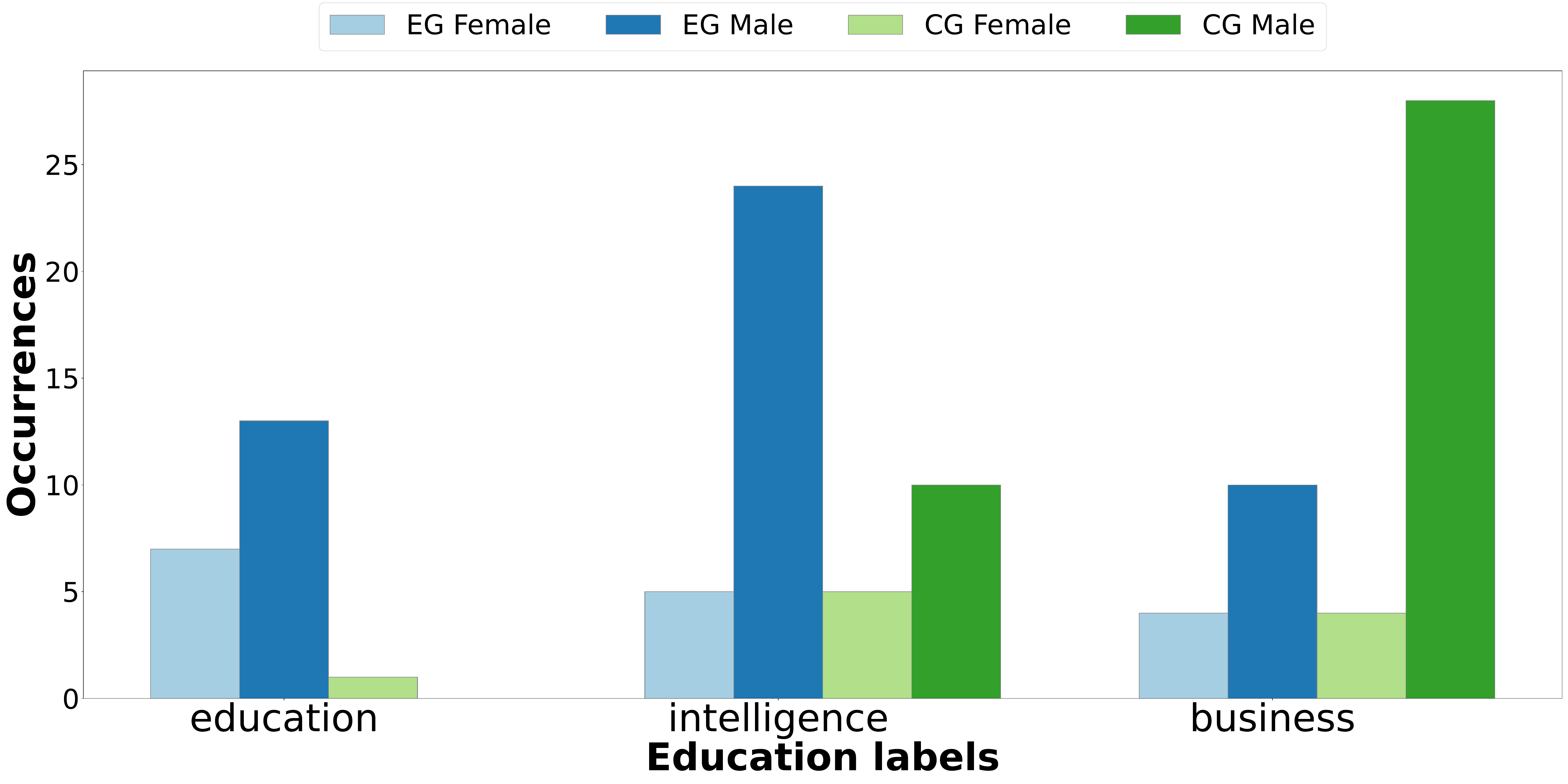}
        \caption{
            Gendered comparison between experimental group and control group.
        }
        \label{fig:education_Gendered_Clarifai}
    \end{subfigure}
    \caption{Comparison regarding \textit{Education} labels assigned by the ClarifAI model.}
\end{figure}

Figure \ref{fig:aesthetic_Clarifai} shows that the EG had a higher number of occurrences than the CG for every concept except the label \textit{Sexy}, although the difference is very small. 
Looking at the gender distinction, Figure \ref{fig:aesthetic_Gendered_Clarifai}, it is noticeable that for both the EG and the CG, women were more likely to be associated with the aesthetic labels than men.
In terms of numbers, 316 vs. 135 labels were assigned to females and males, respectively.
Furthermore, the label \textit{Sexy} is only assigned to those images that were classified as female in the gender recognition task.
The description of this label is linked to the ability to arouse sexual desire or interest, which is only associated with the female gender.
Regarding the labels of the category \textit{Education}, Figure \ref{fig:education_Gendered_Clarifai}, most of the labels were associated with images representing males rather than females.
The overall situation reflects the typical stereotypes associated with gender, with aesthetic labels associated with females and educational labels associated with males.
 
\subsubsection{Person Descriptors}

\begin{figure*}[t]
    \centering
    \begin{subfigure}[b]{0.49\linewidth}
        \includegraphics[width=\linewidth]{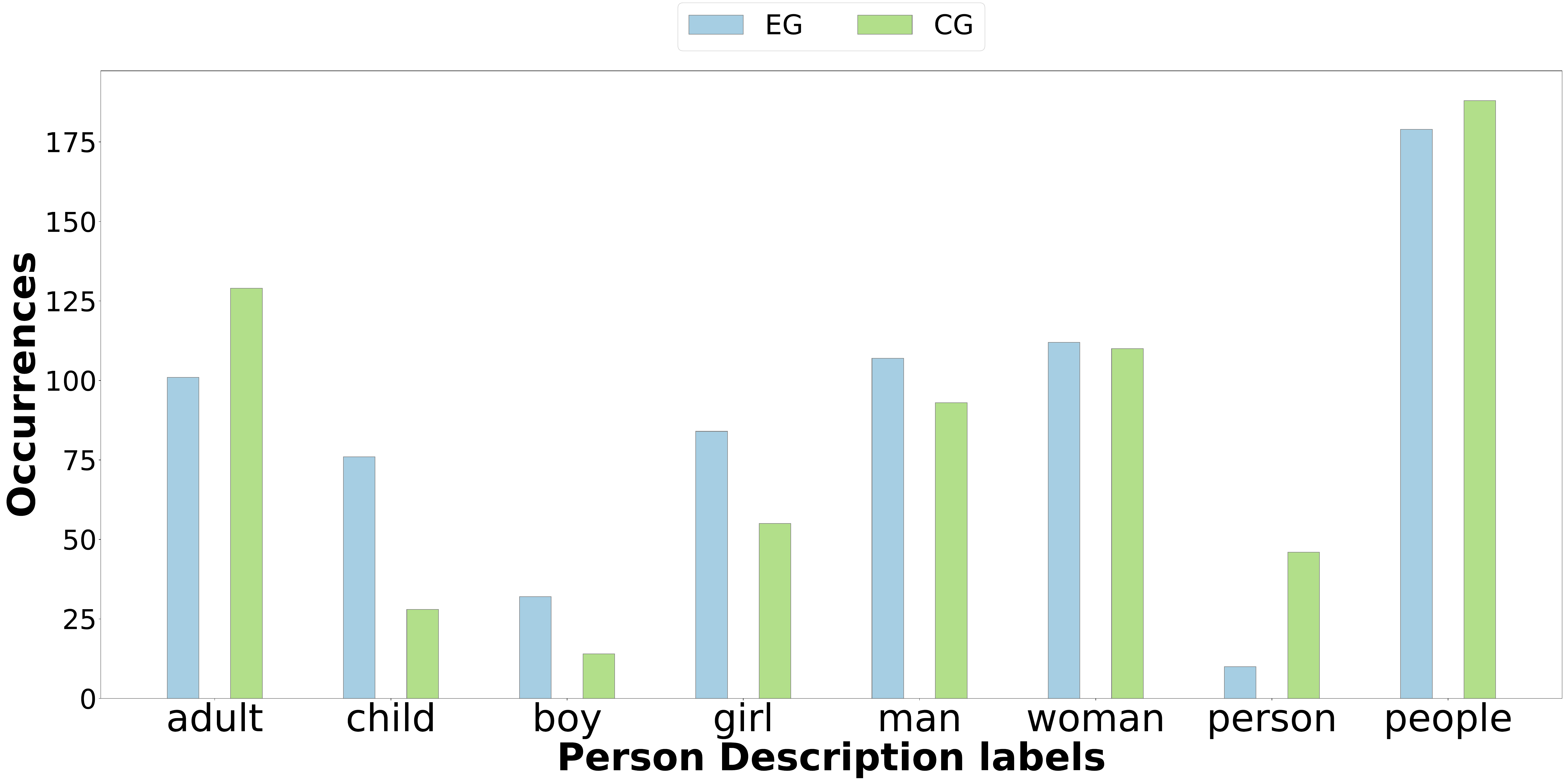}
        \caption{
            Comparison between experimental group and control group regarding \textit{Person Description} labels assigned by the ClarifAI model.
        }
        \label{fig:PersonalDescription_Clarifai}
    \end{subfigure}
    \hfill
    \begin{subfigure}[b]{0.49\linewidth}
        \includegraphics[width=\linewidth]{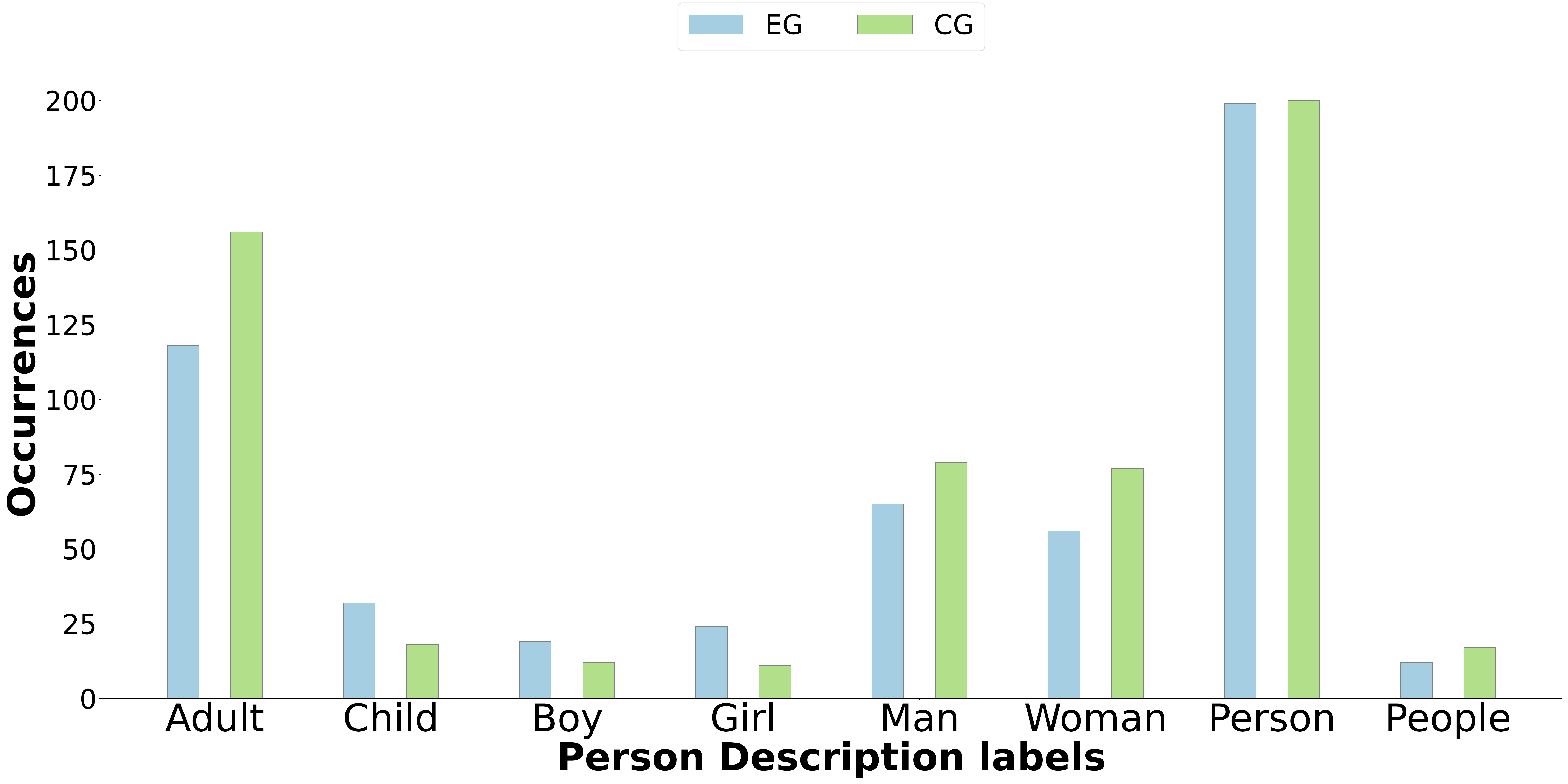}
        \caption{
            Comparison between experimental group and control group regarding \textit{Person Description} labels assigned by the AWSR model.
        }
    \label{fig:PersonDescription_AWS}
    \end{subfigure}
    \\
    \begin{subfigure}[b]{0.48\linewidth}
        \includegraphics[width=\linewidth]{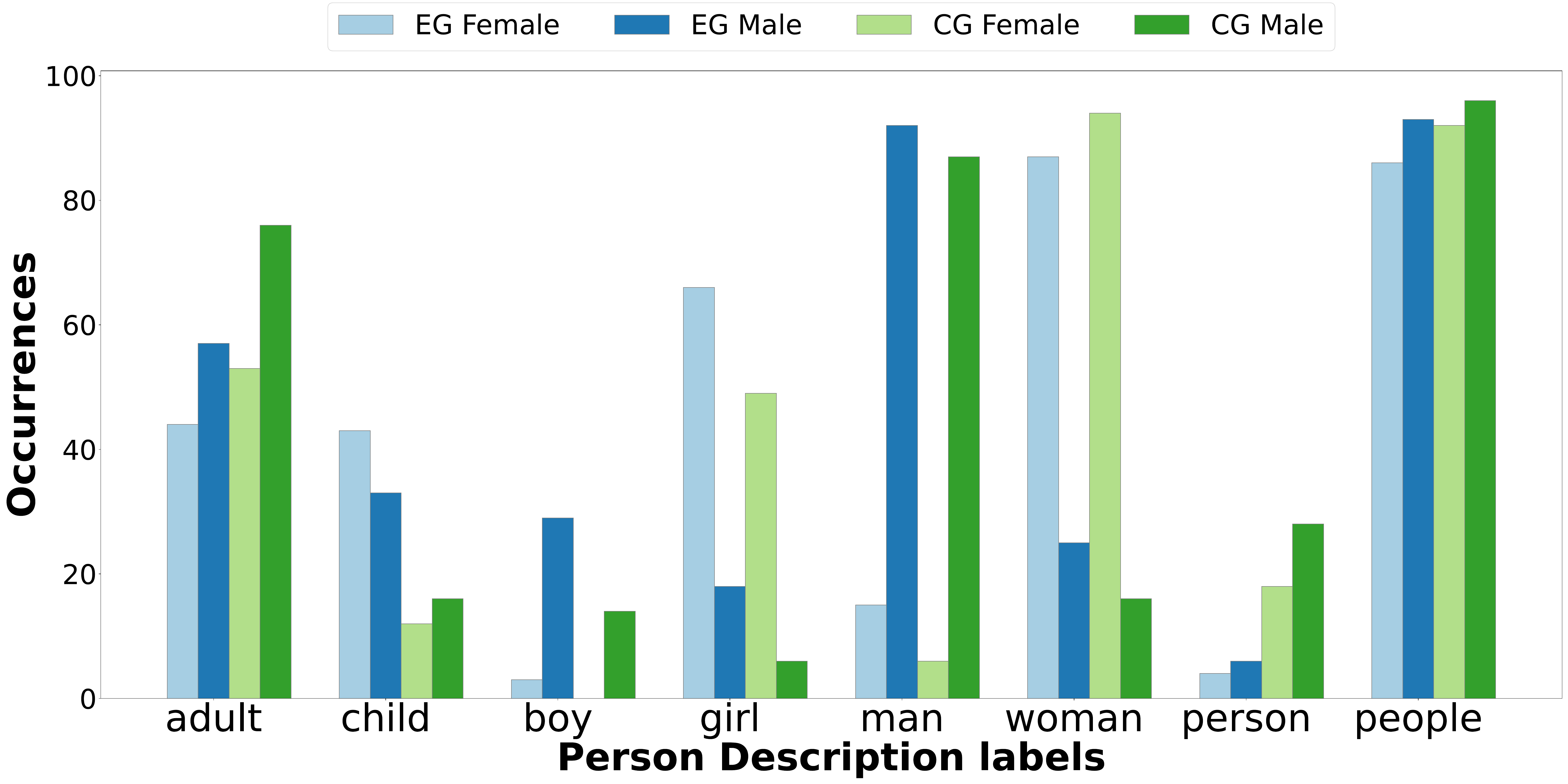}
        \caption{
            Gendered comparison between EG and CG regarding \textit{Person Description} labels assigned by the ClarifAI model.
        }
        \label{fig:PersonDescription_Clarifai_gendered}
    \end{subfigure}
    \hfill
    \begin{subfigure}[b]{0.48\linewidth}
        \includegraphics[width=\linewidth]{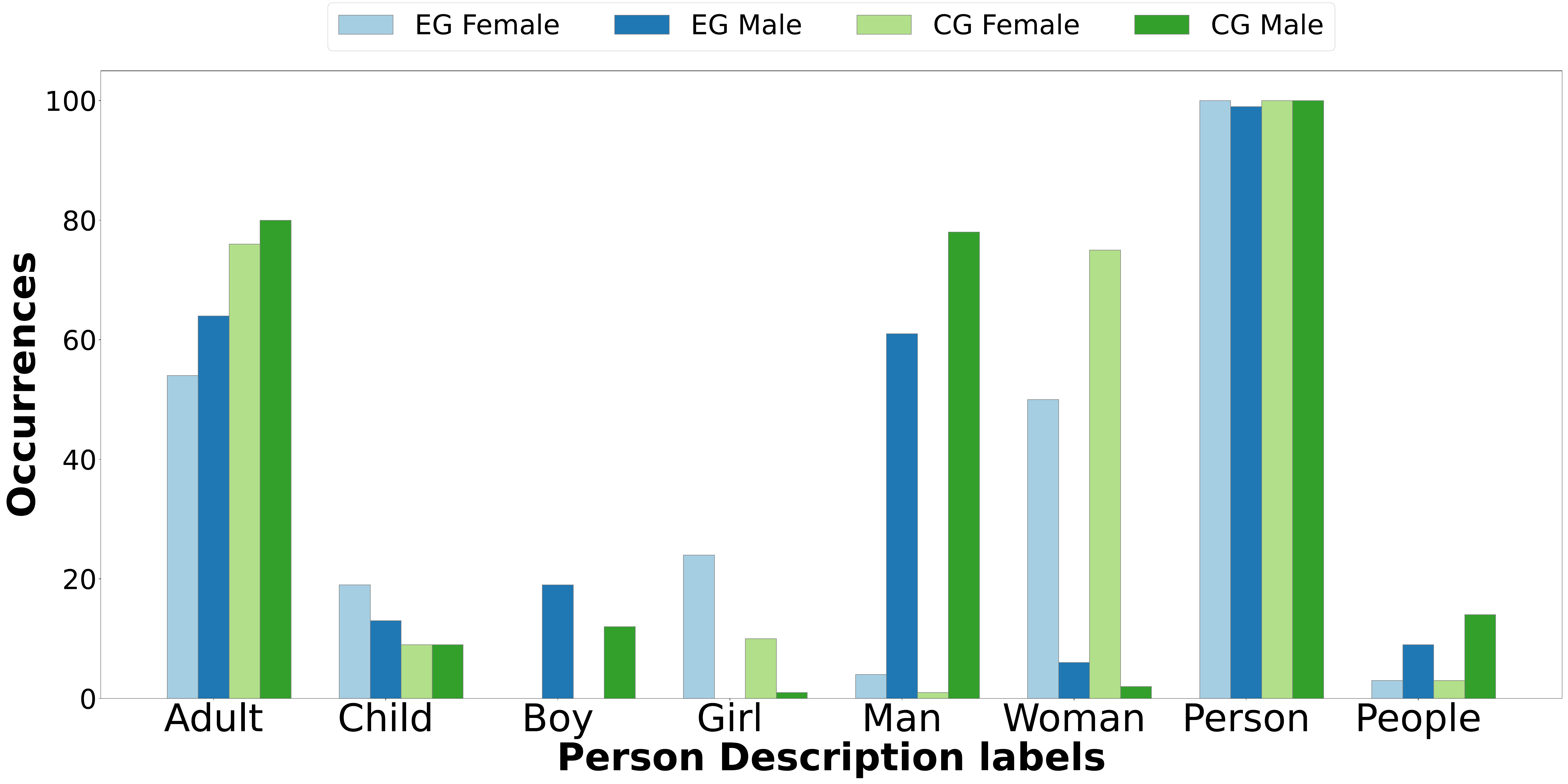}
        \caption{
            Gendered comparison between EG and CG regarding \textit{Person Description} labels assigned by the AWSR model.
        }
        \label{fig:PersonDescription_AWS_gendered}
        \end{subfigure}
        \caption{Comparison regarding \textit{Person Description} labels assigned by the ClarifAI and AWSR models.}
\end{figure*}

\vspace{-0.3cm}
The name \textit{Person description} is taken from one of the predefined categories of the AWSR model.
All the labels are common to both models, so a comparison can be made as shown in Figure \ref{fig:PersonalDescription_Clarifai} and Figure \ref{fig:PersonDescription_AWS}.

Both models assigned the label \textit{Child} more often to the EG than to the CG, although the number of images representing children is quite balanced between the two groups.
The same consideration can be made for the label \textit{Adult}, where the situation is reversed: this result is consistent with the observations from the age prediction task, i.e. the models were more likely to consider a person in the EG as a child rather than an adult.

Each image represents only one person, and according to the definition given by the ClarifAI model for the label \textit{Person} - one human being - and the label \textit{People} - (plural) any group of people (men or women or children) together - the label
the label \textit{Person} is the correct one for each image in the dataset. 
ClarifAI used the correct label for only 56 images out of a total of 400 images, while AWSR used the correct label for 399 images out of a total of 400 images. 

The other labels are gendered in the sense that they refer strictly to one gender rather than the other.
For this reason, it may be useful to look at the Figure \ref{fig:PersonDescription_Clarifai_gendered} and \ref{fig:PersonDescription_AWS_gendered} where a comparison is made of the labels assigned between all four different classes of the dataset: EG male, EG female, CG male, CG female.
All of these gendered labels occurred for both genders, meaning that some images representing females were labelled as \textit{Man} and vice versa.
Some images had both labels \textit{Man} and \textit{Woman} or \textit{Boy} and \textit{Girl}.
Both of these considerations reflect a general confusion in the assignment of these types of labels.

In addition, even if the task of the model is different from gender recognition, gender is always seen as binary, reflecting specific values and beliefs.
Some labels in the \textit{Person Description} category, such as \textit{Male}, \textit{Female}, \textit{Man}, \textit{Woman}, etc., have the power to classify people on the basis of their appearance.
Incorrect gender labelling can have a number of negative consequences for people, especially for groups that are systematically discriminated against. 
It is therefore questionable whether the use of gender labels is appropriate.
\vspace{0.5cm}

\framebox{
    \begin{minipage}[h]{0.95\linewidth}
        The results obtained by the image labelling models do not show significant differences between the EG and the CG, but they show very important differences between the genders. In particular, labels of the \textit{Aesthetics} category are more likely to be associated with female classes than with male classes, and labels of the \textit{Education} category are more likely to be associated with male classes than with female classes.
    \end{minipage}
}


\section{Threats to Validity and Ethical Concerns}\label{sec:threats-validity-limitation}
Most of the limitations and ethical concerns of the study relate to the construction of the dataset and the choice of models.

Starting from the limitations of the dataset, some images are of poor quality.
The resizing of the images to obtain the cropped versions inevitably lowered the quality of about 15\% of the images belonging to the CG. 
The impact of the lower quality of these images on the classification results cannot be excluded, but this problem affected a limited number of images.

As described in Section \ref{sec:Dataset}, some images of the EG did not include information about the year in which they were taken or the age of the person depicted. 
The resulting limitation is an unbalanced comparison between the two groups, EG and CG, in terms of the different number of samples for each age range. 

Since there is no information and no details on which datasets were used for the learning phase of the models, we cannot exclude the possibility that the same images, or some of them, were used in the training process. 
Building a test set with homemade images requires relevant economic and organizational resources.

As the dataset has been created using resources available online, we do not have the explicit consent of the people depicted in the images. 
For this reason, the dataset created is not available to the public and it will remain private.

Another important limitation relates to the \textit{ethnicity} of the people represented.
We carefully constructed the dataset to include people from different backgrounds, so that a variety of ethnicities are included in the dataset.
However, given the difficulties in finding images for the EG, we didn't aim at an equal distribution of different ethnicities, leaving this aspect to future work.

\section{Conclusions}\label{sec:conclusions}

The goal of this study was to understand whether people with Down syndrome may experience problems when their facial image is automatically classified. 
By focusing on this specific group of vulnerable people, we identified a gap in the literature on bias in facial analysis systems and further contributed to the investigation of the inherent limitations of this technology. 

To achieve our goal, we created a test set by collecting facial images already available on the web.
We collected 400 images, 200  faces of people with Down syndrome (experimental group, EG) and 200 faces of people without the syndrome (control group, CG).
We then compared the performance of two commercial face recognition tools.
The results showed that overall the tools performed less well with the EG. 
We also found that: i) the gender prediction showed a higher error rate towards the Down male sample, with an accuracy value of 85\% for both tools; ii) people with Down syndrome were assigned a younger age in relation to their real age; iii) the labels assigned to the experimental group reflected the same gender stereotypes observed in the labels of the control group, in certain cases with a higher frequency.

Involving the most vulnerable populations in the design of facial analysis systems can reduce their operational bias.
Improving the transparency of documentation could also allow for better external scrutiny.
However, we should question the technology itself and its implementation. 
Predicting sensitive characteristics such as gender and age, as well as tagging a person's face with pre-defined labels, could have significant consequences for the lives of those people.
Regardless of the level of accuracy achievable, this technological development may not be socially acceptable. 
This work contributes to this debate by shedding further light on the structural limitations of facial analysis systems.

\section{Future Work}\label{sec:future_work}

The way in which the dataset was constructed was, for the time being, the most feasible way of investigating the research questions.
One of the first improvements in the construction of the dataset is to involve people from the selected communities and ask them to take photographs of themselves, thus obtaining their consent and some valuable information. 
This could be a valid solution to some limitations mentioned above, such as knowing the exact age of each person and getting their explicit consent to be part of the research. 
Furthermore, some problems encountered during the process, such as pose, lighting and quality, can be solved by using appropriate cameras and rules for taking photographs.

Another important improvement concerns ethnicity. 
An idea to construct an equally balanced dataset can be inspired by the \textit{Pilot Parliaments Benchmark} dataset \cite{buolamwiniGenderShadesIntersectional2018}.

In terms of models, it would be interesting to increase the number of models studied. 
This will make it possible to get a broader picture of how FASs are performing in relation to groups that are under-represented.

Further progress could be made on active measures to reduce discrimination against under-represented groups (such as people with Down syndrome).
Since the subgroup performance issues that lead to dangerous discrimination stem most probably from under-representation and unbalanced data, it would be interesting to explore a way to measure data characteristics (such as balance) and provide ad hoc designed labels that provide ethically relevant information. 
Such a tool could be integrated into the AI pipeline to allow developers to be aware of the data issues and take into consideration meaningful countermeasures.
This could be also useful if it is used to publish and disseminate relevant information related to public datasets that are widely used in the AI community, such as those used or mentioned in this article.

\clearpage

%
%
\bibliographystyle{splncs04}
\bibliography{03_frt-ds}
\end{document}